\documentclass{article}
\PassOptionsToPackage{numbers,sort&compress}{natbib}
\usepackage[preprint]{neurips_2026}
\usepackage[utf8]{inputenc}
\usepackage[T1]{fontenc}
\usepackage{graphicx}
\usepackage{amsmath,amssymb}
\usepackage{booktabs,multirow}
\usepackage[dvipsnames,table]{xcolor}
\usepackage{pifont}
\usepackage{tcolorbox}
\usepackage{microtype}
\usepackage{placeins,needspace}
\usepackage[hyphens]{url}
\definecolor{linkblue}{rgb}{0.21,0.49,0.74}
\usepackage[breaklinks,colorlinks,allcolors=linkblue]{hyperref}
\newcommand{\cmark}{\ding{51}}
\newcommand{\xmark}{\textcolor{gray!55}{\ding{55}}}
\newcommand{\pmark}{$\triangle$}
\definecolor{dQual}{HTML}{2F6DB5}
\definecolor{dCam}{HTML}{2E8B57}
\definecolor{dWorld}{HTML}{C98B00}
\definecolor{dDyn}{HTML}{D2691E}
\definecolor{dCons}{HTML}{7B52AB}
\definecolor{dRev}{HTML}{C0392B}
\newtcbox{\mtag}[1][dQual]{on line, arc=3pt, boxrule=0.6pt,
  colback=#1!12, colframe=#1, coltext=#1!80!black,
  left=2.5pt, right=2.5pt, top=0.6pt, bottom=0.6pt, boxsep=0pt,
  fontupper=\bfseries\footnotesize}

\title{WanderBench: Can World Models Explore Dynamic Worlds from Video?}

\author{
Jiheng Li$^{1*}$ \quad
Jiale Wang$^{1*}$ \quad
\textbf{Yuhang Chen}$^{1}$\thanks{Equal contribution.}%
\kern0.4em\thanks{Project leader}\quad
Zai Wang$^{1*}$ \quad Guangyue Liu$^{1}$ \quad \\[4pt] % Shilong Liu$^{3}$ \\[4pt]
$^{1}$Huazhong University of Science and Technology \\[2pt]
\texttt{thisiscyh@gmail.com}
}
\hypersetup{
  pdftitle={WanderBench: Can World Models Explore Dynamic Worlds from Video?},
  pdfauthor={Jiheng Li；; Jiale Wang; Yuhang Chen ；Zai Wang;},
  pdfsubject={Video-conditioned world model evaluation},
  pdfkeywords={WanderBench, WanderLand, world models, video generation}
}

\begin{document}
\maketitle

\begin{abstract}
Video world models conditioned on an input video make it possible to explore
real, recorded dynamic scenes, enabling re-shooting of footage,
free-viewpoint replay, and real-to-sim environments for embodied agents.
Such uses require the generated rollout to stay faithful to the recorded
world, yet existing benchmarks start from a single image or a text prompt,
and video-conditioned generation is evaluated mainly on perceptual quality
and camera following. These evaluations do not directly assess whether exploration preserves subject identity, event progression, and scene consistency upon revisiting a location. We introduce \textbf{WanderBench}, a benchmark for
exploring dynamic worlds from video. It comprises 745 single-shot source
videos of real and AI-generated content in three difficulty tiers
(3--25\,s), paired with camera routes given as keyboard actions, and
evaluates six dimensions with 24 human-validated sub-metrics. Benchmarking
eight models, we find that as clips grow longer, models keep the static
scene and visual quality but lose camera control and the source dynamics,
and that none keeps revisited places consistent. We further propose
\textbf{WanderLand}, a training-free framework that conditions each frame
of a frozen model on the same-instant source and, on revisits, on its own
earlier static content. WanderLand improves five models on every
dimension, by up to 8.3 points in source-dynamics fidelity.
\end{abstract}
\section{Introduction}
\label{sec:intro}
\begin{table}[t]
\centering
\caption{Comparison with representative video-generation and world-model benchmarks.
\textbf{Video}/\textbf{Action}: video / action input. \textbf{Long}: long-horizon evaluation.
The six dimensions follow WanderBench: visual quality (\textbf{Qual.}), camera control (\textbf{Cam.}), source-world fidelity (\textbf{World}), source-dynamics fidelity (\textbf{Dyn.}), spatiotemporal consistency (\textbf{Cons.}), and dynamic revisit consistency (\textbf{Revisit}).
\cmark: explicit design goal; \pmark: partial or indirect coverage; \xmark: not covered.
WanderBench is the only benchmark that conditions on a recorded video and scores exploration against its time-synchronized source events.}
\label{tab:compare}
\small
\setlength{\tabcolsep}{4pt}
\resizebox{\linewidth}{!}{%
\begin{tabular}{l ccc cccccc r}
\toprule
\textbf{Benchmark} & \textbf{Video} & \textbf{Action} & \textbf{Long} & \textbf{Qual.} & \textbf{Cam.} & \textbf{World} & \textbf{Dyn.} & \textbf{Cons.} & \textbf{Revisit} & \textbf{\#Cases} \\
\midrule
VBench~\cite{huang2023vbench,huang2025vbench++,zheng2025vbench} & \xmark & \xmark & \xmark & \cmark & \xmark & \xmark & \xmark & \pmark & \xmark & 946 \\
WorldScore~\cite{duan2024worldscore}           & \xmark & \cmark & \cmark & \cmark & \cmark & \pmark & \pmark & \cmark & \xmark & 3,000 \\
WorldModelBench~\cite{wang2024worldmodelbench} & \xmark & \cmark & \xmark & \xmark & \pmark & \pmark & \pmark & \xmark & \xmark & 350 \\
MIND~\cite{ye2026mind}                         & \xmark & \cmark & \cmark & \cmark & \cmark & \pmark & \xmark & \cmark & \pmark & 250 \\
Omni-WorldBench~\cite{wu2026omni}              & \xmark & \cmark & \pmark & \cmark & \cmark & \xmark & \pmark & \cmark & \xmark & 1,068 \\
WorldMark~\cite{xu2026worldmark}               & \xmark & \cmark & \cmark & \cmark & \cmark & \xmark & \xmark & \cmark & \pmark & 500 \\
WBench~\cite{ying2026wbench}                   & \xmark & \cmark & \cmark & \cmark & \cmark & \pmark & \pmark & \cmark & \pmark & 289 \\
R2M-Bench~\cite{gu2026r2m}                     & \xmark & \cmark & \cmark & \xmark & \xmark & \xmark & \xmark & \xmark & \pmark & 300 \\
WorldMemBench~\cite{zhou2026worldmembench}     & \xmark & \cmark & \cmark & \cmark & \cmark & \xmark & \pmark & \xmark & \cmark & 260 \\
\midrule
\rowcolor{blue!6}
\textbf{WanderBench (Ours)}                    & \cmark & \cmark & \cmark & \cmark & \cmark & \cmark & \cmark & \cmark & \cmark & 745 \\
\bottomrule
\end{tabular}}
\end{table}
Recent advances in video generation have steadily improved visual quality, motion realism, and temporal coherence, and the scope of the field has broadened beyond producing fixed-length clips to interactive world models, in which users continuously issue camera or action controls during generation to explore an unfolding visual world~\cite{kong20253d,huang2025vid2world,gao2026advancing,sun2025worldplay,team2026alayaworld,feng2024thematrix,he2025matrix,wang2026matrix}. Such models offer new tools for game creation~\cite{li2025hunyuan,tang2025hunyuan,bruce2024genie,parkerholder2024genie2,genie3,zhang2025matrixgame,he2025matrix,wang2026matrix}, virtual production~\cite{rahimi2025generative,team2025hunyuanworld,mao2025yume,mao2026yume1,team2026advancing}, and robot policy learning~\cite{xiao2025world,yang2023learning,zhu2025irasim}. However, most world models start from a single image or a text prompt, so the scene dynamics they produce are entirely imagined by the model and do not correspond to events that actually took place.

Video-conditioned world models open up a new possibility: given a real recorded video of a dynamic scene, users can re-observe the scenes and events it captures along new camera trajectories~\cite{team2026inspatio,xu2026wonder,song2026world}. This capability directly supports re-framing real footage and free-viewpoint replay, and offers a way to build simulation environments for embodied agents from real videos. Unlike exploration from a single image, these applications impose stricter requirements on the generated results: the model must not only synthesize plausible novel views but also remain faithful to the world recorded in the source video~\cite{bai2025recammaster,yu2025trajectorycrafter}.

Despite these advances, it remains unclear whether current
models can faithfully explore the dynamic world captured
in a source video. Existing benchmarks evaluate visual
quality~\cite{huang2023vbench,zheng2025vbench}, interactive control~\cite{wu2026omni,ying2026wbench,liang2026worldlens}, and world consistency~\cite{xu2026worldmark,ye2026mind},
including memory across revisits~\cite{gu2026r2m,zhou2026worldmembench}. However, success on these dimensions does not necessarily
imply fidelity to the recorded event. A model may generate
realistic views and follow the requested camera trajectory
while altering the source action or its timing. It may
also produce seemingly consistent revisits by retaining
an outdated state of a dynamic subject.
This leaves a central question: can models use the source
video as evidence of an evolving world, preserving its
events under viewpoint changes, integrating information
across frames, and maintaining scene consistency during
revisits while respecting the event's temporal progression?

To address this gap, we introduce \textbf{WanderBench}, a benchmark for evaluating video-conditioned world exploration. Each test case pairs a single-shot source video of a dynamic scene with a camera route specified by discrete keyboard actions, which determines how the events in the source video should be observed from changing viewpoints. WanderBench contains 745 source videos spanning real-world footage such as dance, film, and natural scenes, together with AI-generated videos in realistic, anime, and artistic styles, organized into three difficulty tiers from 3 to 25 seconds. We provide random exploration and out-and-back camera routes to examine continuous viewpoint changes and revisits to previously observed regions, respectively. The evaluation comprises 24 sub-metrics across six complementary dimensions. By incorporating source-video references, event timing, and visibility, WanderBench assesses whether stable scene structure and subject identities are preserved and whether dynamic content evolves in accordance with the source event, providing a systematic evaluation of faithful reproduction and sustained exploration of dynamic worlds under camera control.

Benchmarking eight video-conditioned world models on WanderBench reveals two consistent failure modes. First, as clips grow longer, models keep the static scene and their visual quality but lose camera control and the source dynamics: the output drifts from the commanded view and falls out of sync with the source events. Second, no model remembers what it has generated, so a revisited place is redrawn and changes appearance. Models limited to a few seconds of generation cannot even reach the longer tiers.

Beyond evaluation, we introduce \textbf{WanderLand}, a training-free
inference-time framework for video-conditioned world exploration.
WanderLand aims to preserve source-world identity and event progression
while maintaining consistency throughout camera-controlled exploration
and revisits. To this end, we develop
an evidence-ordered conditioning scheme that rebuilds only the camera and source render a model already reads: every frame is anchored to the source camera of the same instant and rendered from the same-instant source frame, and on revisits, the regions the source cannot cover are filled with the static content the model itself generated earlier. It
operates on pretrained video world models without updating their
parameters, using only each model's own geometry. We evaluate WanderLand on WanderBench across
five video world models, demonstrating
consistent gains across all six dimensions, with the largest improvements in source-dynamics fidelity (up to $+8.3$) and dynamic revisit consistency (up to $+4.8$), the two dimensions targeted by the failure modes above.

Our main contributions are:
\begin{itemize}
    \item We introduce \textbf{WanderBench}, a benchmark for world exploration from dynamic video inputs. It comprises 745 single-shot source videos organized into three difficulty tiers, with random camera exploration and long-horizon out-and-back revisit tasks across all tiers, providing a unified setting for evaluating exploration of source-video worlds from novel viewpoints.

    \item We establish an evaluation framework comprising six complementary dimensions and 24 sub-metrics. By incorporating source-video references, event timing, and visibility, it systematically assesses stable properties that should persist and dynamic content that should evolve with the source event. Benchmarking eight models with it, we identify loss of camera control and source dynamics over long horizons, and inconsistent revisits, as shared failure modes.

    \item We propose \textbf{WanderLand}, a training-free framework for enhancing existing video world models. Built upon these models, WanderLand uses evidence-ordered conditioning, which draws on the same-instant source and the model's own earlier static content, to improve source-content preservation and dynamic-event reproduction. We evaluate it on WanderBench across five applicable models and observe gains on all six dimensions, most notably in source-dynamics fidelity and dynamic revisit consistency.
\end{itemize}
\section{Related Work}
\label{sec:related}
\label{sec:composition}

\subsection{Video World model}
Early world models explored robotic manipulation and autonomous driving, including UniSim~\cite{yang2023learning}, IRASim~\cite{zhu2025irasim}, GAIA-1~\cite{hu2023gaia}, and Vista~\cite{gao2024vista}. With advances in video generation, world models can continuously generate subsequent frames from visual observations and user actions, enabling interactive exploration of generated environments. YUME~\cite{mao2025yume,mao2026yume1} introduces natural language into world event control, while HY-World~\cite{hy-worldHYworld20Multimodal2026} and LingBot-World~\cite{lingbot-world} explore world generation that combines real-time interaction with long-term consistency. Hunyuan-GameCraft~\cite{li2025hunyuan} and Matrix-Game~\cite{zhang2025matrixgame,he2025matrix,wang2026matrix} support keyboard-and-mouse-driven interactive generation. AlayaWorld~\cite{team2026alayaworld} further supports sustained world exploration through compressed temporal history and geometrically aligned spatial memory.

Further work uses source videos as references for scene appearance and dynamic content, allowing users to re-observe the events they contain along new camera trajectories. ReCamMaster~\cite{bai2025recammaster} jointly conditions on a source video and a target camera trajectory to generate novel views while preserving the source video's appearance consistency and dynamic synchronization. WorldForge~\cite{song2025worldforge,song2026worldforge} improves trajectory-controlled video generation through reference-fusion correction and denoising guidance. InSpatio-World~\cite{team2026inspatio} supports spatiotemporal exploration through world states anchored to a reference video; Wonder~\cite{xu2026wonder} jointly designs visual camera control, sparse memory, and distillation strategies to support real-time world generation conditioned on images and videos; and World in World~\cite{song2026world} converts source observations, target-view projections, and historical information into unified visual evidence, enabling training-free control through the pretrained model's native self-attention. WanderBench addresses these requirements by systematically evaluating video world models' ability to controllably explore dynamic worlds from source videos. Models must not only generate plausible novel views but also preserve the scene and subject identities in the source video, maintain the temporal progression of the original event, and remain consistent with the current event state when revisiting previously observed regions.

\subsection{World Models Evaluation}

Existing benchmarks broadly fall into two categories: non-interactive video evaluation and world model evaluation. VBench~\cite{huang2023vbench,zheng2025vbench,huang2025vbench++}, EvalCrafter~\cite{liu2024evalcrafter}, and VideoPhy~\cite{bansal2025videophy,bansal2026videophy} do not support action inputs or interaction, focusing primarily on video quality, text alignment, or physical commonsense. In contrast, WorldScore~\cite{duan2024worldscore} evaluates camera-trajectory-conditioned generation, WorldModelBench~\cite{wang2024worldmodelbench} studies decision-oriented world-model quality, WorldArena~\cite{shang2026worldarena} targets embodied agents in closed domains, MIND~\cite{ye2026mind} probes closed-loop memory consistency, Omni-WorldBench~\cite{wu2026omni} focuses on causal interaction, WorldLens~\cite{worldlens2025} targets autonomous driving, and WorldMark~\cite{xu2026worldmark} measures navigation consistency. WBench further introduces a multi-turn evaluation framework for open-domain interactive world models and R2M-Bench~\cite{gu2026r2m} evaluates consistency during long-horizon revisits in interactive video world models. Other efforts examine complementary aspects~\cite{han2025videobench,liu2024fetv,gu2025phyworldbench,cai2025mmgr,upadhyay2026worldbench,lu2025_4dworldbench,lian2024worldsimbench,yue2025ewmbench,zhang2025worldinworld,tang2025hunyuan,zhou2026drivinggen}.

These studies provide a foundation for evaluating generation quality, interactive control, and world consistency. However, in video-conditioned settings, evaluation must also determine whether the generated results remain faithful to the dynamic world recorded in the source video: models should preserve scene and subject identities while reproducing the actions and temporal progression of the original event. WanderBench addresses this problem by incorporating source-video references, event timing, and visibility into 24 sub-metrics across six complementary dimensions, systematically evaluating source-world fidelity, source-dynamics fidelity, and consistency during exploration and revisits under camera control.

\section{WanderBench: Can World Models Explore Dynamic Worlds from Video?}
\label{sec:wanderbench}

\begin{figure}[t]
  \centering
  \includegraphics[width=\linewidth]{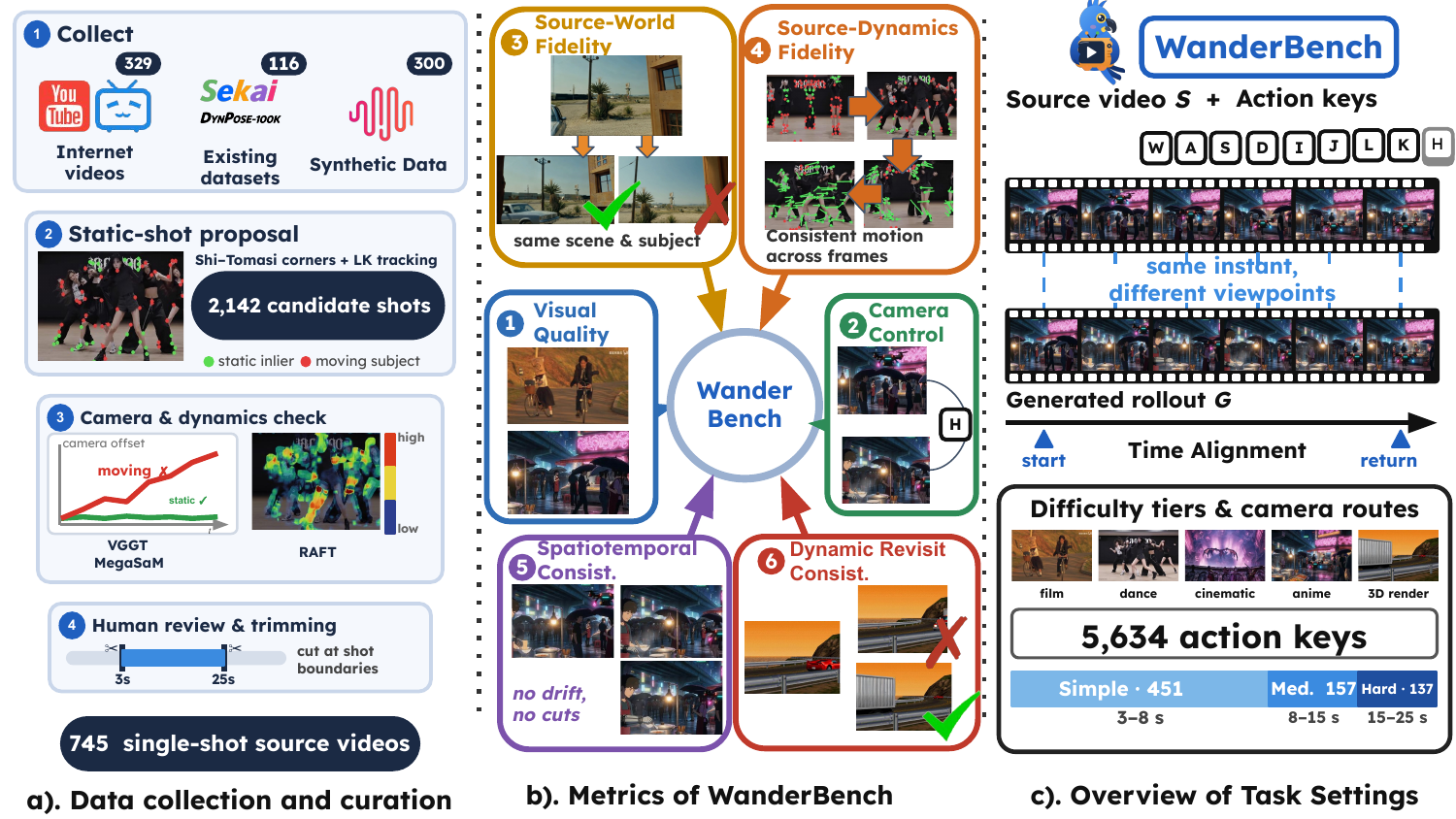}
  \caption{\textbf{Overview of WanderBench.}
  (a) Data collection and curation.
  (b) Six evaluation dimensions with 24 sub-metrics.
  (c) Task setting: given a source video and action keys, a model re-renders the scene from novel viewpoints while staying synchronized with the source.}
  \label{fig:overview}
\end{figure}
\textbf{Data Composition.}
WanderBench comprises 745 source videos, each a single continuous shot whose camera remains static or near-static, so that $C(t)=\mathbf{I}$ denotes the original view.
329 are real-world or rendered videos collected from the Internet (Bilibili and YouTube), covering dance, film, daily life, stage performances, and other scenes; 116 are drawn from existing video datasets~\cite{huang2025vipe,li2025sekai,rockwell2025dynamic}; and the remaining 300 are generated with MiniMax, spanning 13 realistic looks, 12 anime looks, and 48 artistic.

\textbf{Clip Screening.}
Candidate clips are screened in two stages.
(1) Shot proposal. For Internet videos, static shots are automatically proposed by following Shi--Tomasi~\cite{shi1994good} corners with a pyramidal Lucas--Kanade~\cite{lucas1981iterative} optical flow at 10 fps and fitting a RANSAC~\cite{fischler1981random} similarity transform.
(2) Camera and dynamics verification. For each candidate, we reconstruct the camera trajectory with VGGT~\cite{wang2025vggt}, cross-checked with MegaSaM~\cite{li2024megasam}, and discard clips with noticeable camera rotation or translation. We further check scene dynamics with RAFT~\cite{teed2020raft} optical flow.

\textbf{Difficulty Tiers.}
WanderBench is organized into three tiers of increasing difficulty:
(1) Simple (451 clips) contains short clips with slow, mild subject motion;
(2) Medium (157 clips) features longer clips with higher visual quality and richer lighting variation;
and (3) Hard (137 clips) contains the longest clips, with the highest visual quality, large-amplitude subject motion, substantial background changes, and complex illumination.
Across the benchmark, camera-compensated motion extends from a median of 0.029 to 0.12 image diagonals, and MUSIQ imaging quality spans from 22.9 to 74.8.
To verify the single-shot filtering, four annotators independently labeled all 2{,}142 candidate clips as single-shot or not.
The annotators agree with each other at Fleiss' $\kappa=0.93$, yielding the final 745 clips.

\subsection{Evaluation Suite}
\label{sec:metrics}

WanderBench decomposes evaluation into six complementary dimensions comprising 24 sub-metrics, detailed in the Appendix.
Most of our metrics compare the output with a time-synchronized source reference: each output frame $G(t)$ and the source frame $S(\tau(t))$ depict the same instant from two viewpoints, so every frame has a reference without any temporal annotation.
All sub-metric scores are rescaled to $[0,100]$.

\noindent\textbf{$\bullet$ Visual Quality.}
We adopt \mtag[dQual]{V.1} Aesthetic Quality, \mtag[dQual]{V.2} Imaging Quality, \mtag[dQual]{V.3} Temporal Flickering, and \mtag[dQual]{V.4} Motion Smoothness from VBench~\cite{huang2023vbench}, together with \mtag[dQual]{V.5} HPSv3 Preference~\cite{hpsv3} and \mtag[dQual]{V.6} Visual Plausibility scored by the PAVRM model of WBench~\cite{ying2026wbench}. Since the task is to re-render an input video, all scores are normalized by those of the source video.

\noindent\textbf{$\bullet$ Camera Control.}
\mtag[dCam]{C.1} Synchronized Camera Following reconstructs the source and generated videos separately with VGGT~\cite{wang2025vggt} and compares the realized camera motion with the commanded one at one-second intervals.

\noindent\textbf{$\bullet$ Source-World Fidelity.}
\mtag[dWorld]{W.1} Scene Adherence and \mtag[dWorld]{W.2} Subject Adherence use a VLM and source subject tracks, respectively, to check whether the scene facts and subjects of the source are preserved. \mtag[dWorld]{W.3} Synchronized View Fidelity reprojects the source frame of the same instant into the realized view and compares it with the output on co-visible regions.

\noindent\textbf{$\bullet$ Source-Dynamics Fidelity.}
\mtag[dDyn]{D.1} Motion Timeline Agreement compares the camera-compensated motion curves of each subject between the source and the output. \mtag[dDyn]{D.2} Event Adherence localizes the source events in the output. \mtag[dDyn]{D.3} Temporal Synchronization finds the best-matching source time for each output frame and measures the proportion that falls within a tolerance of the correct time.

\noindent\textbf{$\bullet$ Spatiotemporal Consistency.}
\mtag[dCons]{S.1} Subject Consistency and \mtag[dCons]{S.2} Background Consistency penalize only the drift of subjects and static regions beyond the changes in the source. \mtag[dCons]{S.3} Segment Continuity penalizes hard cuts absent from the source. \mtag[dCons]{S.4} Static Geometric Consistency measures the reprojection consistency of static regions with Depth Anything 3~\cite{lin2025depth}.

\noindent\textbf{$\bullet$ Dynamic Revisit Consistency.}
\mtag[dRev]{R.1}--\mtag[dRev]{R.5} are based on DreamSim similarity (R.1), its pose-gated variant (R.2), place descriptors (R.3), geometric correspondence (R.4), and object appearance (R.5), respectively. \mtag[dRev]{R.6} Object Semantic Persistence checks the objects that should appear in the source at the return time. \mtag[dRev]{R.7} Revisit Event Currency measures whether the returning view shows the current state of a dynamic subject.
\section{Experiments}

\subsection{Evaluated Models and Protocol}
\label{sec:exp_setup}

We evaluate eight video-conditioned world models that take a source video and a target camera trajectory, divided into two groups by generation length: \textbf{(1) full-length} models, which render the whole clip in one pass, including InSpatio-World~1.5~\cite{team2026inspatio}, World-in-World~\cite{song2026world}, and WorldForge~v2~\cite{song2026worldforge};
and \textbf{(2) frame-limited} models, which produce at most 81 frames (49 for TrajectoryCrafter), including ReCamMaster~\cite{bai2025recammaster}, CameraAnything~\cite{cameraanything2026}, UniWorld-View~\cite{zhou2026uniworld}, WorldForge~v1~\cite{song2025worldforge}, and TrajectoryCrafter~\cite{yu2025trajectorycrafter}.
All models receive the same source video and the same action sequence.
Each action sequence is converted into per-frame camera poses relative to the source camera and passed through the model's native conditioning interface at its native resolution and frame rate; models with a different frame rate receive the same actions resampled to their own time grid.
Frame-limited models process only the opening segment of each clip up to their maximum length (81 frames, or 49 for TrajectoryCrafter), conditioned on the corresponding prefix of the full trajectory, and are scored on that segment only.
For scoring, (1) all metrics operate on a time grid in seconds;
(2) scores are averaged over items, videos, and then models;
and (3) a dimension score is the mean of its sub-metrics, and the overall score is the equal-weight mean of the six dimensions.

% Single-column NeurIPS layout: 9 pt text, no whole-table scaling.
\definecolor{cIns}{HTML}{E6EFFB}
\definecolor{cWiW}{HTML}{E7F4EA}
\definecolor{cWF}{HTML}{FCEFE2}
\definecolor{cLim}{HTML}{EDEDED}
\newcommand{\bb}[1]{\textbf{\textcolor{black}{#1}}}

\begin{table}[t]
\centering
\caption{\textbf{Results on WanderBench by difficulty tier.} Overall is the equal-weight mean of the six dimensions. Bold marks the best score within each tier. Rows of the same model share a colour; grey rows are frame-limited models, which are evaluated only on Simple and only on the frames they can generate.}
\label{tab:tiers}
\small
\setlength{\tabcolsep}{2pt}
\renewcommand{\arraystretch}{1.15}
\begin{tabular*}{\linewidth}{@{\extracolsep{\fill}}lccccccc@{}}
\toprule
Model & \shortstack{Visual\\Quality} & \shortstack{Camera\\Control}
& \shortstack{Source-World\\Fidelity} & \shortstack{Source-Dyn.\\Fidelity}
& \shortstack{Spatiotemp.\\Consist.} & \shortstack{Dyn. Revisit\\Consist.} & Overall \\
\midrule
\multicolumn{8}{@{}l}{\textbf{Simple (3--8\,s), 451 clips}} \\
\multicolumn{8}{@{}l}{\textit{Full-length}} \\
\rowcolor{cIns} InSpatio-World 1.5 & 74.9 & 75.4 & 87.5 & \bb{65.8} & 83.4 & 54.5 & 73.6 \\
\rowcolor{cWiW} World-in-World     & 84.0 & 44.4 & \bb{90.5} & \bb{65.8} & 83.3 & 53.1 & 70.2 \\
\rowcolor{cWF} WorldForge v2      & \bb{85.8} & 44.8 & 88.1 & 60.9 & \bb{85.9} & 50.0 & 69.3 \\
\rowcolor{cLim}\multicolumn{8}{@{}l}{\textit{Frame-limited} (scored on the first $\leq$81 frames; 49 for TrajectoryCrafter)} \\
\rowcolor{cLim} UniWorld-View      & 68.7 & \bb{85.1} & 89.6 & 61.2 & 84.5 & \bb{68.6} & \bb{76.3} \\
\rowcolor{cLim} CameraAnything     & 74.3 & 66.5 & 85.5 & 65.4 & 82.8 & 46.0 & 70.1 \\
\rowcolor{cLim} TrajectoryCrafter  & 72.1 & 74.8 & 75.7 & 58.1 & 80.7 & 56.5 & 69.7 \\
\rowcolor{cLim} WorldForge v1      & 78.7 & 60.5 & 84.4 & 53.0 & 82.7 & 51.8 & 68.5 \\
\rowcolor{cLim} ReCamMaster        & 73.3 & 40.8 & 87.2 & 62.0 & 83.0 & 32.8 & 63.2 \\
\midrule
\multicolumn{8}{@{}l}{\textbf{Medium (8--15\,s), 157 clips}} \\
\rowcolor{cIns} InSpatio-World 1.5 & 73.2 & \bb{70.6} & 87.3 & \bb{55.2} & 83.2 & \bb{52.7} & \bb{70.4} \\
\rowcolor{cWiW} World-in-World     & \bb{82.9} & 41.2 & \bb{89.9} & 49.9 & 83.0 & 51.4 & 66.4 \\
\rowcolor{cWF} WorldForge v2      & 79.9 & 43.6 & 86.5 & 41.3 & \bb{84.8} & 40.8 & 62.8 \\
\multicolumn{8}{@{}l}{\color{gray}\textit{Frame-limited models: not applicable, clips exceed their generation length.}} \\
\midrule
\multicolumn{8}{@{}l}{\textbf{Hard (15--25\,s), 137 clips}} \\
\rowcolor{cIns} InSpatio-World 1.5 & 71.6 & \bb{49.0} & 81.7 & 40.2 & \bb{80.5} & \bb{50.5} & \bb{62.3} \\
\rowcolor{cWiW} World-in-World     & \bb{78.4} & 34.7 & \bb{86.5} & \bb{44.8} & 78.6 & 50.1 & 62.2 \\
\rowcolor{cWF} WorldForge v2      & 74.0 & 10.4 & 82.0 & 26.6 & 79.3 & 20.5 & 48.8 \\
\multicolumn{8}{@{}l}{\color{gray}\textit{Frame-limited models: not applicable, clips exceed their generation length.}} \\
\bottomrule
\end{tabular*}
\end{table}

\subsection{Findings}
\label{sec:findings}
\textbf{(1) Visual quality hides exploration failures.} From Simple to Hard, the three full-length models lose little in Visual Quality and Spatiotemporal Consistency. Over the same range, they lose much more in Camera Control, Source-Dynamics Fidelity, and Dynamic Revisit Consistency. A rollout can therefore look clean and stable while following neither the command nor the source events, which per-frame quality metrics alone would not reveal.

\textbf{(2) Models keep the static world but lose the dynamic one.} Source-World Fidelity is the highest-scoring dimension on every tier, so scenes and subjects are largely preserved. Source-Dynamics Fidelity, in contrast, falls to 37.2 on Hard, meaning the models keep what is in the scene but not when things happen in it.

\textbf{(3) Visual quality and camera control trade off.} Across the eight models on Simple, Visual Quality and Camera Control are negatively correlated. WorldForge has the highest Visual Quality but only 44.8 in Camera Control, while UniWorld-View has the best Camera Control and the lowest Visual Quality. Faithfully executing a camera move appears to cost image quality in current models, and ranking them by quality alone favors models that move the camera less.

\textbf{(4) Short horizons hide the differences that long horizons reveal.} On Simple, the frame-limited UniWorld-View obtains the highest overall score, and the overall scores of the full-length models span only 4.3 points. On Hard, this span widens to 13.5 points, and WorldForge collapses to 10.4 in Camera Control. Their strong short-horizon scores do not show whether they can explore.

\textbf{(5) Model hardly remembers what it has generated.} Dynamic Revisit Consistency is the weakest dimension overall: no model exceeds 68.6, the full-length models stay between 50.0 and 54.5 even on Simple, and WorldForge drops to 20.5 on Hard. Across models, Dynamic Revisit Consistency correlates with Camera Control, which suggests that a model must first reach the revisit pose before consistency there can be achieved.

\begin{figure}[p]
  \centering
  \includegraphics[width=\textwidth,height=0.95\textheight,keepaspectratio]{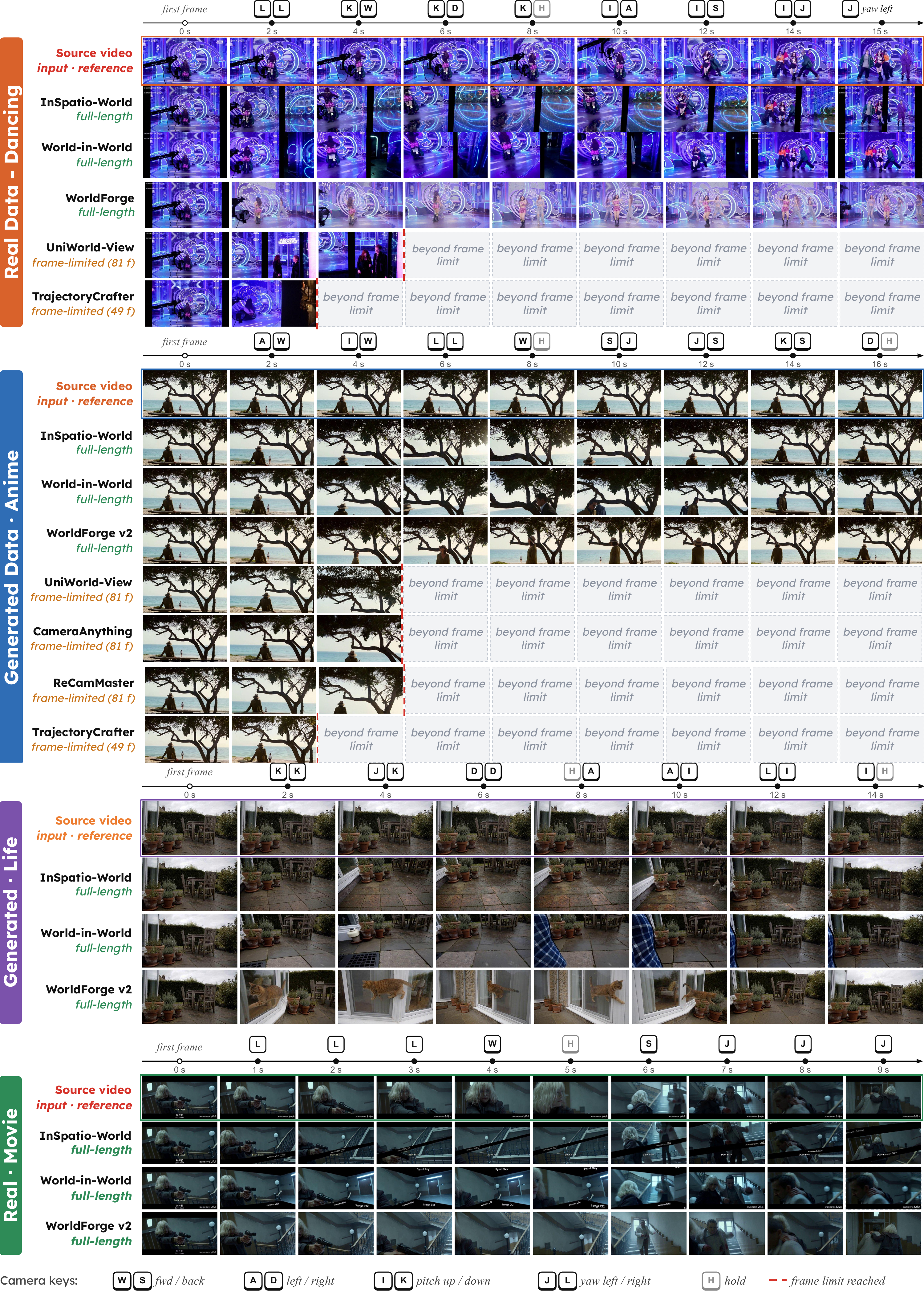}
  \caption{\textbf{Qualitative results on WanderBench.}
  Each block shows one source clip (left tag: data category). Time runs from left to right; the keys above each column are the camera actions held since the previous column, and \textsc{H} denotes hold. The first row of each block is the source video at the same instant, followed by the outputs of each model. Grey cells mark frames beyond a model's generation length, and the red dashed line marks where its frame limit is reached.}
  \label{fig:qualitative}
\end{figure}

\section{WanderLand: A Training-Free Enhancement of Exploration in Video World Models}
\label{sec:wanderland}

To explore how source-video evidence can help address the failure modes identified by WanderBench, we introduce WanderLand, a training-free enhancement for existing video world models. WanderLand constructs conditioning inputs by prioritizing time-aligned source observations and using previously generated static content to fill regions not covered by the source.
\begin{figure}[htbp]
  \centering
  \includegraphics[width=\linewidth]{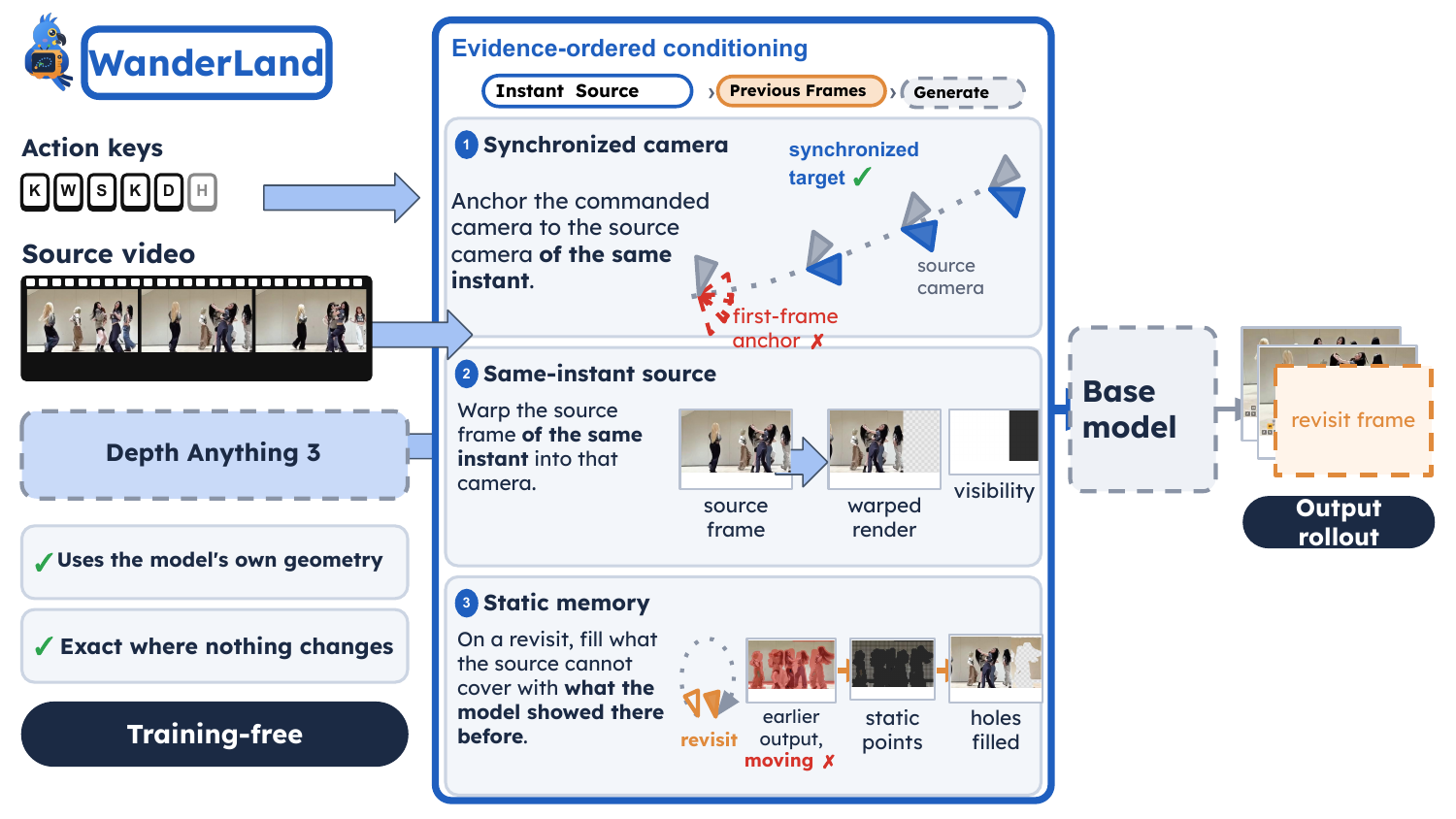}
  \caption{\textbf{Overview of WanderLand.} WanderLand wraps a frozen video world model and leaves its weights, sampler, and seed unchanged. It only rebuilds the camera and source render the model already takes as input, taking evidence for every frame in a fixed order (top): the latent source first, then previous frames, and generation only where neither exists.
  All geometry comes from the model's own front-end. The frozen model then generates the output rollout.}
  \label{fig:wanderland}
\end{figure}

\subsection{Methods}
For each output frame, WanderLand takes evidence in a fixed order, in three steps.

\textbf{(1) Synchronized camera.} The model's own geometry front-end estimates the depth and camera of every source frame, and the commanded camera is then placed relative to the source camera of the same instant rather than to that of the first frame.

\textbf{(2) Same-instant source.} The model's own point-cloud renderer warps the source frame of the same instant into this camera. As a result, the render covers more of the view, and moving subjects appear at the correct time.

\textbf{(3) Static memory.} When the camera returns to a place shown earlier, we lift the model's nearest earlier outputs to 3D points using Depth Anything 3~\cite{lin2025depth} depth aligned to the render depth, and use them to fill the regions the source video cannot cover. Pixels whose source content changes over time, measured by a camera-compensated color difference, are discarded. Moving subjects are therefore never frozen, and all visible motion still comes from the source video.

\subsection{Properties} 
WanderLand requires no training and introduces no additional weights. It relies only on the model's own geometry module and renderer and never uses WanderBench's evaluation tools. Wherever the evidence does not change, it reproduces the original output exactly, and its only extra cost is one additional generation when the camera revisits a place.

\subsection{WanderLand Results}
\label{sec:wanderland_results}

% Single-column NeurIPS layout: 9 pt text, compact wrapped headers.
\newcommand{\up}[1]{\textcolor[HTML]{2E8B57}{+#1}}
\newcommand{\dn}[1]{\textcolor[HTML]{C0392B}{$-$#1}}

\begin{table}[t]
\centering
\caption{\textbf{Change brought by WanderLand} on each dimension (score with WanderLand minus score without). World-in-World and WorldForge are not evaluated, since both are themselves training-free frameworks built on pretrained world models.}
\label{tab:wanderland}
\small
\setlength{\tabcolsep}{2pt}
\renewcommand{\arraystretch}{1.15}
\begin{tabular*}{\linewidth}{@{\extracolsep{\fill}}lcccccc@{}}
\toprule
Model & \shortstack{Visual\\Quality} & \shortstack{Camera\\Control}
& \shortstack{Source-World\\Fidelity} & \shortstack{Source-\\Dynamics\\Fidelity}
& \shortstack{Spatiotemporal\\Consistency} & \shortstack{Dynamic\\Revisit\\Consistency} \\
\midrule
InSpatio-World 1.5 & \up{4.0} & \up{2.2} & \up{2.4} & \up{8.3} & \up{1.8} & \up{2.4} \\
UniWorld-View      & \up{1.2} & \up{1.0} & \up{1.3} & \up{1.9} & \up{1.4} & \up{2.6} \\
TrajectoryCrafter  & \up{0.7} & \up{0.9} & \up{0.5} & \up{0.7} & \up{0.4} & \up{3.0} \\
ReCamMaster        & \up{2.6} & \up{2.8} & \up{0.4} & \up{1.4} & \up{1.0} & \up{3.7} \\
CameraAnything     & \up{2.4} & \up{1.3} & \up{0.5} & 0.0      & \up{1.3} & \up{4.8} \\
\bottomrule
\end{tabular*}
\end{table}

We apply WanderLand to five video world models and report the change on each dimension in Tab.~\ref{tab:wanderland}. World-in-World and WorldForge are not included, since both are themselves training-free frameworks built on pretrained world models and do not expose the inputs that WanderLand rebuilds.

 \textbf{(1) Consistent gains without trade-offs.} WanderLand improves most backbone on every dimension and degrades none, so the better conditioning comes at no cost to visual quality or camera control.
 
\textbf{(2) Revisits benefit on every model.} Dynamic Revisit Consistency rises by 2.4--4.8 points across all five backbones, as the holes at a revisit are filled with content the model has already shown instead of being redrawn. 

\textbf{(3) Long horizons benefit the most.} The largest single gain is on the full-length InSpatio-World~1.5, whose Source-Dynamics Fidelity rises by 8.3 points: over long rollouts, anchoring every frame to the source at the same instant keeps the generated events in sync with the recorded ones.

\begin{figure}[htbp]
  \centering
  \includegraphics[width=\textwidth]{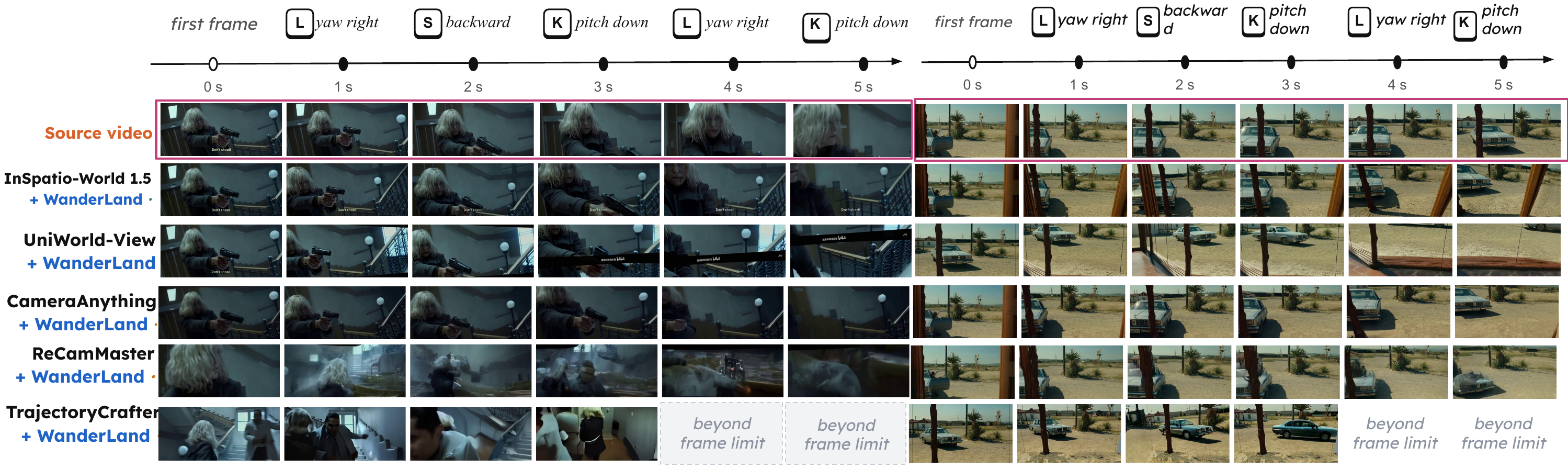}
  \caption{\textbf{Qualitative results of WanderLand.}
  Two clips from WanderBench (left: an indoor action scene; right: an outdoor desert scene).
  Each column shows one frame per second, and the key above it is the camera action held during the preceding second.
  The first row is the source video; the other rows are the outputs of five base models with WanderLand applied.
  TrajectoryCrafter is limited to 49 frames, so its output ends after 3\,s.
  WanderLand keeps the content that is visible in the source and stays consistent under the same key sequence, without retraining any base model.}
  \label{fig:wanderland_examples}
\end{figure}

\section{Discussion}
\label{sec:discussion}

WanderBench moves video world model evaluation from a single view to time-aligned multiple views: each explored trajectory is a new view of the same dynamic scene, synchronized with the source.
We expect models that do well here to produce time-aligned multi-view videos from a single input, which could enable end-to-end generation of full-scene 4D Gaussian Splatting from one video.
Current models still fall short on temporal synchronization and long-range revisits, and we hope WanderBench offers a concrete target for closing this gap.
\Needspace{11\baselineskip}
\section{Conclusion}
\label{sec:conclusion}

We introduced WanderBench, a benchmark for exploring dynamic worlds from video, with 745 single-shot source videos in three difficulty tiers and 24 sub-metrics over six dimensions. Current world models preserve static scenes and visual quality but lose camera control, source dynamics, and revisit consistency over longer horizons. Our training-free WanderLand improves five models by conditioning each frame on the same-instant source and the model's own earlier content, though it requires camera and source-render inputs. We hope WanderBench drives world models that stay faithful to the recorded world as they explore it.

\FloatBarrier
{\small
\setlength{\bibsep}{2pt}
\bibliographystyle{plainnat}
\bibliography{main}
}
\end{document}